\PassOptionsToPackage{hyphens}{url}
\pdfoutput=1

\documentclass[11pt]{article}

\usepackage{acl}

\usepackage{times}
\usepackage{latexsym}

\usepackage[T1]{fontenc}

\usepackage[utf8]{inputenc}

\usepackage{microtype}

\usepackage{inconsolata}
\usepackage{url}
\usepackage{booktabs}
\usepackage{multirow}
\usepackage{graphicx}
\usepackage{amsfonts}
\usepackage{amssymb}
\usepackage{bbding}
\usepackage{dashrule}
\usepackage{subfigure}
\usepackage{colortbl}
\usepackage{amsmath}
\usepackage[ruled,linesnumbered]{algorithm2e}

\title{LLM-DA: Data Augmentation via Large Language Models for \\Few-Shot Named Entity Recognition}

\author{
    \bf{\normalsize
    Junjie Ye$^{1}$,\ \ Nuo Xu$^{1}$,\ \ Yikun Wang$^{1}$,\ \ Jie Zhou$^{2}$,}\\
    \bf{\normalsize Qi Zhang$^{1}$\thanks{Corresponding authors.},\ \ Tao Gui$^{3*}$,\ \ Xuanjing Huang$^1$} \\ 
  {$^1$ \normalsize School of Computer Science, Fudan University} \\
  {$^2$ \normalsize School of Computer Science and Technology, East China Normal University}\\
  {$^3$ \normalsize Institute of Modern Languages and Linguistics, Fudan University} \\
  \texttt{\normalsize jjye23@m.fudan.edu.cn}\\
  \texttt{\normalsize \{qz, tgui\}@fudan.edu.cn} \\
  }

\begin{document}
\maketitle
\begin{abstract}
Despite the impressive capabilities of large language models (LLMs), their performance on information extraction tasks is still not entirely satisfactory. However, their remarkable rewriting capabilities and extensive world knowledge offer valuable insights to improve these tasks. In this paper, we propose \emph{LLM-DA}, a novel data augmentation technique based on LLMs for the few-shot NER task. To overcome the limitations of existing data augmentation methods that compromise semantic integrity and address the uncertainty inherent in LLM-generated text, we leverage the distinctive characteristics of the NER task by augmenting the original data at both the contextual and entity levels. Our approach involves employing 14 contextual rewriting strategies, designing entity replacements of the same type, and incorporating noise injection to enhance robustness. Extensive experiments demonstrate the effectiveness of our approach in enhancing NER model performance with limited data. Furthermore, additional analyses provide further evidence supporting the assertion that the quality of the data we generate surpasses that of other existing methods.
\end{abstract}

\section{Introduction}
\label{sec:intro}

Recently, a series of large language models (LLMs) exemplified by ChatGPT~\footnote{\url{https://openai.com/chatgpt}} have garnered significant attention~\cite{GPT, BLOOM, LLaMA}. These models have undergone extensive training using diverse datasets from various domains and have showcased remarkable competitiveness across various natural language processing (NLP) tasks.

\begin{figure}[!t]
    \centering
    \includegraphics[width=\linewidth]{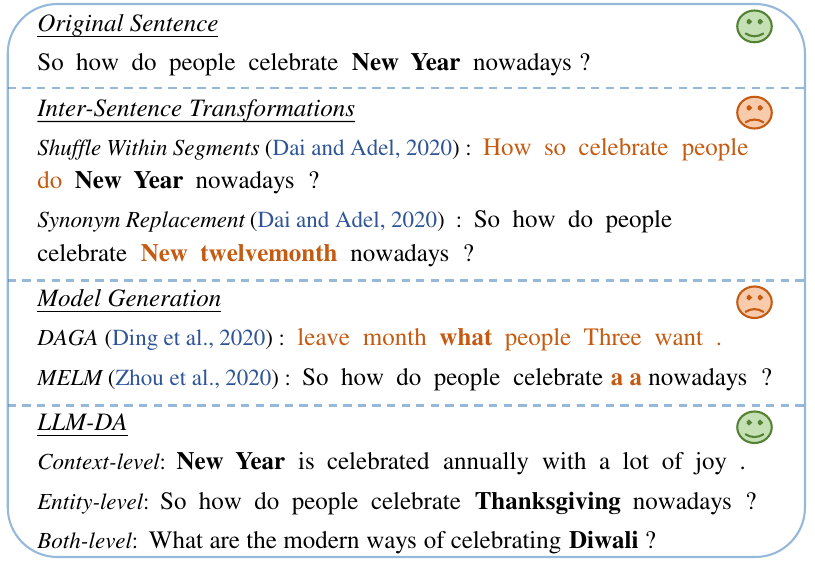}
    \caption{Examples of various data augmentation methods applied to the same original sentence. Entities (``event'') within the sentences are \textbf{bolded}, while any ungrammatical or semantically incoherent parts are highlighted in {\color[RGB]{197,90,17}red}.}
    \label{fig:augmented-data}
\end{figure}

Extensive research has demonstrated that despite the Generative Pre-trained Transformer (GPT)~\cite{GPT} family of models being the most capable LLMs currently available, they still struggle to achieve satisfactory performance on information extraction tasks. 
For instance, the GPT-3.5 series models achieve a micro-F1 score ranging from only 20 to 60 points on the named entity recognition (NER) task~\cite{GPT-chen, GPT-ye}. 
Additionally, training LLMs specifically tailored to a particular domain poses challenges due to the substantial computational resources required and limited adoption~\cite{InstructUIE}. However, the remarkable rewriting ability and vast world knowledge~\cite{World-knowledge} of LLMs offer promising insights for improving the handling of such tasks.

Given the challenges and expenses associated with annotating each token in the NER training data, the availability of annotated samples is limited. Therefore, we focus primarily on addressing the NER problem in scenarios with a scarcity of training data.
The prevailing approach involves utilizing data augmentation techniques to generate additional samples, which can be categorized into two main types. The first category entails applying transformation operations to a given sentence, such as entity substitution, synonym replacement, and shuffle with segments~\cite{LSMS}. The second category involves training a base model, such as LSTM~\cite{LSTM} or RoBERTa~\cite{RoBERTa}, and leveraging its generative capabilities to produce new data.

However, existing approaches often compromise the semantic integrity of the sentences, as illustrated in Figure \ref{fig:augmented-data}.
It is evident that these approaches either neglect the syntax of the sentences (e.g., shuffle within segments and DAGA~\cite{DAGA}) or overlook the semantics of the context (e.g., synonym replacement and MELM~\cite{MELM}), hindering the model's ability to analyze the sentences accurately.
Considering that the NER task requires understanding the syntactic structure and semantic information of each token in a sentence to accurately identify entities and their corresponding types, current data augmentation techniques may generate sentences that deviate from factual accuracy and are not conducive to effective model training.

To tackle the issues, we propose \emph{LLM-DA}, a novel data augmentation technique based on LLMs. 
LLM-DA utilizes the text generation capabilities of LLMs to produce semantically coherent sentences adhering to actual expression conventions. However, a drawback lies in the high uncertainty associated with the content generated by LLMs, which presents a challenge in handling the generated data. 
To impose necessary constraints on the generated content while preserving the diversity of the augmented data, we apply data augmentation at both the context and entity levels. This approach also aligns with the inherent characteristics of the NER task, which primarily focuses on identifying entities and their categories.

To enrich \textbf{context-level augmentation}, we present a diverse set of 14 contextual rewriting strategies across four dimensions: sentence length, vocabulary usage, subordinate clause incorporation, and presentation style. This expanded repertoire aims to enhance diversity while maintaining semantic coherence.
For \textbf{entity-level augmentation}, we leverage the extensive world knowledge of LLMs to substitute entities in a sentence with others of the same type. This enables us to generate diverse entities that adhere closely to the sentence semantics, surpassing the constraints of the training data.
To further enhance the diversity of the augmented data, we employ a \textbf{both-level augmentation} approach, combining both augmentation methods.
In addition, to facilitate the model's adaptability to variations in the input,
we incorporate noise injection~\cite{Noise} by generating sentences with random spelling errors.

We compare the results obtained from four variants of our proposed method against four other baseline approaches. The experimental findings demonstrate that our approach
significantly improves the model's performance in the few-shot NER task. 
Furthermore, through a comprehensive analysis of the data generated by various data augmentation methods, we observe that LLM-DA consistently produces data of higher quality compared to existing approaches.

To summarize, the key contributions of our paper are listed as follows:
\begin{itemize}
    \item
    We introduce a novel data augmentation method that leverages the world knowledge and rewriting capability of LLMs. Experimental results demonstrate the significant performance improvement achieved by our approach to the NER task, particularly in scenarios with limited training samples.
    \item 
    We augment the existing sentences at both the contextual and entity levels. To ensure the diversity and high quality of the generated data, we employ 14 different contextual rewriting strategies and introduce appropriate noise during the data augmentation process.
    \item
    By conducting a comprehensive analysis of the generated data, we illustrate the consistent superiority of our approach in producing high-quality sentences when compared to existing methods. Furthermore, the augmented data achieve a harmonious equilibrium between diversity and controllability.
\end{itemize}

\begin{figure*}[!t]
    \centering
    \includegraphics[width=\textwidth]{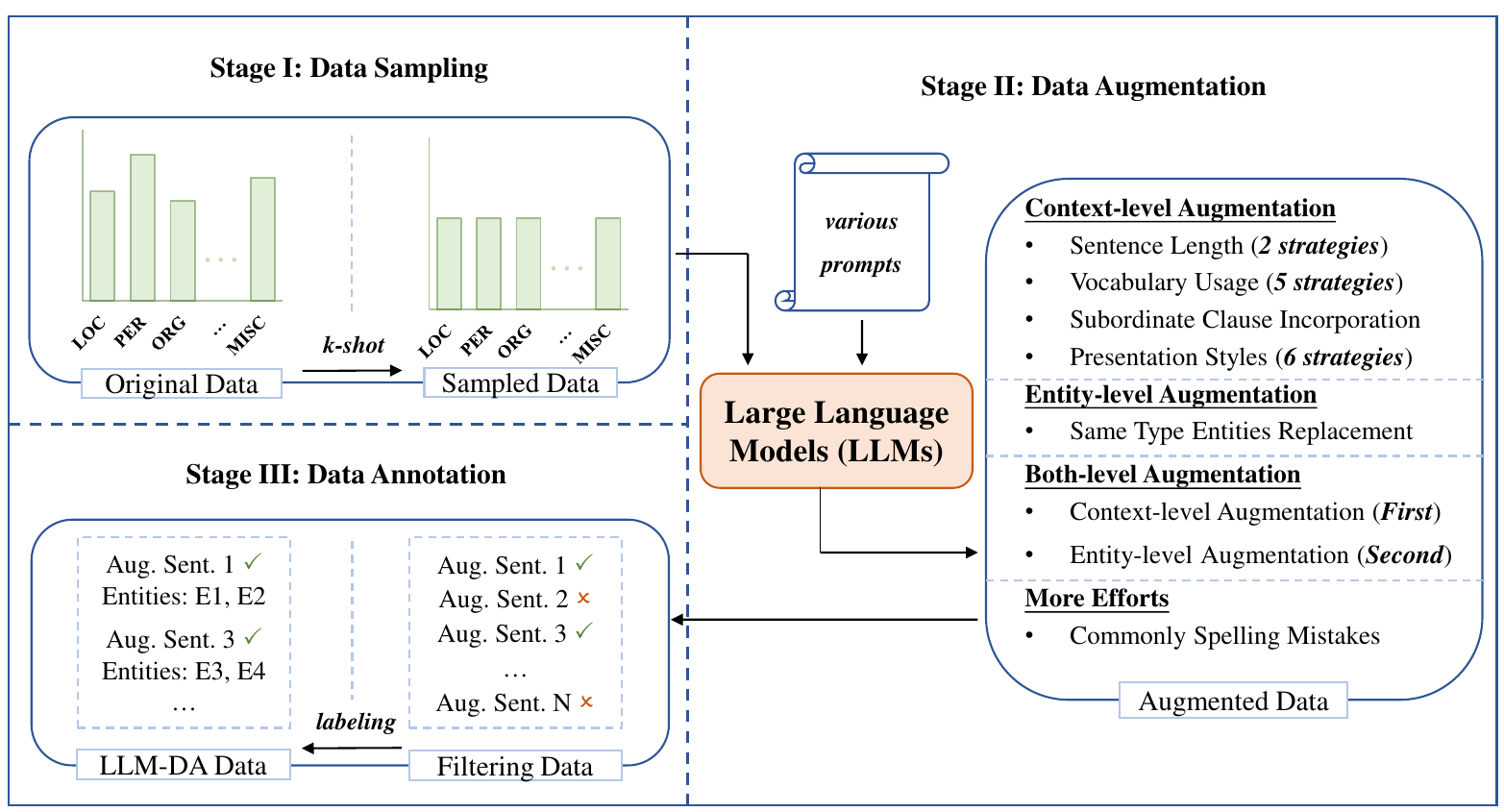}
    \caption{An illustration of LLM-DA. LLM-DA comprises three stages: data sampling from the original dataset, generation of augmented data across four dimensions using LLMs with specially designed prompts, and finally, filtering the augmented data and 
    data annotation to obtain the final valid dataset.}
    \label{fig:LLM-DA}
\end{figure*}

\section{Proposed Method}

LLM-DA is a data augmentation technique primarily relying on LLMs, leveraging their world knowledge and rewriting capabilities to extend the limited labeled data and acquire a substantial volume of high-quality data. Figure \ref{fig:LLM-DA} illustrates the three stages of LLM-DA: data sampling, data augmentation, and data annotation. These phases will be elaborated on in the following sections.

\subsection{Data Sampling}
\begin{algorithm}[!t]
    \caption{Greedy Algorithm for Data Sampling}
    \label{algorithm:sampling}
    \KwIn{Number of shot $k$, labeled training set $\mathrm{D}$ with a label set $\mathcal{Y}$.}
    \KwOut{The sampled training set $\mathrm{D_{smp}}$.}
    $\mathrm{D_{smp}} \gets \phi$ \tcp{Initialize the sampled training set.} 
    
    \For{$\forall y \in \mathcal{Y}$}{
    {Count}[$y$] $\gets 0$    
    } 

    $\mathrm{Shuffle~D}$

    \For{$(X,Y) \in \mathrm{D}$}{
    $add \gets \mathrm{True}$

        \For{$y \in \mathcal{Y}$}{
        $\mathrm{Count_{tmp}}[y] \gets$ the mention number of $y$ in $(X,Y)$

        \If{$\mathrm{Count}[y]+\mathrm{Count_{tmp}}[y]>1.25 \times k$}{$add \gets \mathrm{False}$}
        }

        \If{$add==\mathrm{True}$}{
        $\mathrm{D_{smp}} \gets \mathrm{D_{smp}} \cup (X,Y)$

        \For{$y \in \mathcal{Y}$}{
        $\mathrm{Count}[y] \gets \mathrm{Count}[y] + \mathrm{Count_{tmp}}[y]$
        }
        }
        \If{$\forall$ $y \in \mathcal{Y}$, $\mathrm{Count}[y]>=k$}{$\mathrm{break}$ \tcp{Finish sampling.}}
    }  
\end{algorithm}

Given a labeled training set $\mathrm{D}=\{(X_i, Y_i)\}_{i=1}^{\vert \mathrm{D} \vert}$, where $\vert \mathrm{D} \vert$ represents the number of samples in $\mathrm{D}$, our approach involves utilizing a greedy algorithm (Algorithm \ref{algorithm:sampling}) to sample $k$-shot examples. In the context of the NER task, a standard $k$-shot sample refers to obtaining exactly $k$ occurrences of entities for each category in the sample~\cite{SNN}. However, meeting this strict requirement can be challenging, especially when dealing with data that exhibits an imbalanced label distribution. To address this challenge, we draw inspiration from the work of \citet{FEW-NERD} and introduce a relaxation to the criterion. This relaxation permits a maximum of $1.25k$ occurrences of entities per category in the $k$-shot scenario.
Our method takes into account the inherent bias in the distribution of entity categories in the real world while striving to maintain the $k$-shot setting as closely as possible. This approach enables a better capture of the true distribution of real-world data.

\subsection{Data Augmentation}
Since LLMs do not require parameter fine-tuning with data, the careful design of appropriate prompts becomes crucial to fully leverage their capabilities.
Drawing on established principled guidance~\footnote{\url{https://help.openai.com/en/articles/6654000-best-practices-for-prompt-engineering-with-openai-api}} and considering the distinctive nature of the NER task, which involves identifying entities and their types within a given sentence, we construct specialized prompts for augmenting the data at both the context and entity levels.~\footnote{The specific prompts can be found in Appendix \ref{appendix:prompts}.}

\paragraph{Context-level Augmentation} 
Previous work by \citet{MELM} has shown that simple context replacement offers limited improvement.
To diversify the text, we've integrated the rewriting method across four distinct dimensions, drawing insights from prior linguistic studies~\cite{Language}. Our goal is to generate 14 sentence variations for a given input sentence and its associated entities. These variations encompass diverse aspects in terms of sentence length (i.e., long and short), lexical usage (i.e., advance words, adverbs, adjectives, prepositions, and conjunctions), incorporation of subordinate clauses, and presentation styles (i.e., news, spoken language, magazines, fictions, Wikipedia, and movie reviews). Throughout this process, we obtain a set of new sentences with rewritten contexts while preserving the entities from the original sentences.

\paragraph{Entity-level Augmentation}
We instruct the model to substitute entities in a given sentence with other entities of the same type while maintaining the original sentence structure. Unlike existing entity replacement methods~\cite{LSMS, MELM}, our approach leverages the world knowledge embedded within LLMs, allowing the generation of entities beyond those present in the training set. Furthermore, to enhance data diversity, we generate as many sentences as possible while adhering to the token length limitations of the model. This strategy substantially broadens the range of entities and contributes to the overall richness of the augmented data.

\paragraph{Both-level Augmentation}
Building upon the augmentations made at the context and entity levels, we naturally combine these two approaches within a given sentence. However, when both levels are simultaneously rewritten, the inherent uncertainty of the generated content often leads to new sentences that deviate significantly from the original, posing challenges for label handling. To address this, we adopt a two-stage strategy that first performs context-level augmentation and then follows up with entity-level augmentation based on the already generated contextually augmented sentences. By employing this strategy, we achieve effective data diversification while maintaining meaningful connections between the newly generated sentences and the original ones.

Furthermore, to improve the robustness of the downstream model, we instruct LLMs to randomly substitute certain words in the original sentences with common spelling errors. To ensure that the model derives benefits from the introduced noise without being excessively affected by its adverse effects, we generate a maximum of one augmented sentence with noise for each original sentence.

\subsection{Data Annotation}

While we have introduced various constraints in the prompts to guide the output of LLMs, it is important to acknowledge that generative models may still produce content that does not align with our requirements. To overcome this, we implement a straightforward filtering process for the generated data.
Specifically, for context-level augmented data, we discard sentences in which the model retains entities other than given entities. Similarly, for other types of augmented data, we remove sentences in which the model modifies entities that are not given entities. For the filtered sentences, we assign the same labels as the original sentences. This means that the retained or replaced entities are labeled with their original entity types, while the remaining tokens are labeled as non-entity types.

\section{Experimental Setups}
\subsection{Task Formulation}
We primarily evaluate our approach by comparing its performance to existing methods on the few-shot NER task. 
In the NER task, we are given a candidate entity label $\mathrm{E}$ and a sequence of tokens $x = (x_1, x_2, ..., x_n)$. The model's objective is to predict the entity label $y = (y_1, y_2, ... y_n)$, where each $y_i \in \mathrm{E}$ corresponds to a token in $x$.
The $k$-shot NER task refers to a scenario where each entity in the training set occurs approximately $k$ times, which may not provide sufficient training data for the model.

\begin{table}[!t]
    \centering
    \resizebox{\linewidth}{!}{%
    \begin{tabular}{lcccccc}
    \toprule
    {Datasets} & {\# Class} & {\# 5-shot} & {\# 10-shot} & {\# 15-shot} & {\# 20-shot} & {\# Test} \\
    \midrule
    CoNLL'03 & 4 & 10 & 20 & 30 & 40 & 3.7k \\
    OntoNotes 5.0 & 11 & 50 & 100 & 140 & 190 & 8.3k \\
    MIT Movie & 12 & 45 & 90 & 135 & 180 & 2.4k\\
    FEW-NERD  & 66 & 250 & 435 & 620 & 860 & 37.6k \\
    \bottomrule
    \end{tabular}
    }
    \caption{Statistics for datasets. 
    The ``$k$-shot'' column represents the number of data sampled in that scenario.
    }
    \label{datasets}
\end{table}

\subsection{Dataset}

We evaluate our method comprehensively using four NER datasets that span various domains and exhibit distinct levels of granularity. These datasets include CoNLL'03 \cite{conll} from the news domain, OntoNotes 5.0 \cite{ontonotes} from the general domain, MIT-Movie \cite{Mit-movie} from the review domain, and FEW-NERD \cite{FEW-NERD} from the general domain as well. Detailed information regarding each dataset can be found in Table~\ref{datasets}.

\subsection{Baselines}
\paragraph{Data Augmentation Methods}
To evaluate the effectiveness of our approach, we compare it against four baselines in our experiments:
1) \textbf{Gold} employs the original sampled data as training samples to train the NER model;
2) \textbf{LSMS}~\cite{LSMS} utilizes four sentence transformations to generate augmented data, including label-wise token replacement, synonym replacement, mention replacement, and shuffle within segments;
3) \textbf{DAGA}~\cite{DAGA} trains an LSTM model using the sequence with linearized labels and utilizes this model to generate new data for training;
and 4) \textbf{MELM}~\cite{MELM} trains a RoBERTa model and employs it to generate augmented data by replacing entities in the sequence.
At the same time, we categorize our methods into four distinct categories:
1) \textbf{LLM-DA (Context)} employs context-level augmented data as training data;
2) \textbf{LLM-DA (Entity)} utilizes entity-level augmented data along with noise injection data as training data;
3) \textbf{LLM-DA (Both)} incorporates both-level augmented data as training data;
and 4) \textbf{LLM-DA (All)} amalgamates LLM-DA (Context), LLM-DA (Entity), and LLM-DA (Both) data together for training purposes.

\paragraph{Classification Models}
To conduct a thorough performance comparison of our proposed approach with existing data augmentation methods for the few-shot NER task, we utilize two established pre-trained masked language models, subsequently fine-tuning them for evaluation purposes:
1) \textbf{BERT}~\cite{BERT} is a large-scale pre-trained model based on the Transformer \cite{transformers} encoder architecture and is widely utilized in various NLP tasks;
and 2) \textbf{RoBERTa} is an improved iteration of BERT and is renowned as one of the state-of-the-art pre-trained classification models in the field.

\begin{table}[!t]
    \centering
    \resizebox{\linewidth}{!}{
    \begin{tabular}{lcccc}
    \toprule
         Methods&  CoNLL'03&  OntoNotes 5.0&  MIT-Movie& FEW-NERD\\ \midrule
         LLM-DA (Context)&  45$\times$&  20$\times$&  30$\times$& 10$\times$\\
         LLM-DA (Entity)&  20$\times$&  18$\times$&  20$\times$& 15$\times$\\
         LLM-DA (Both)&  40$\times$&  12$\times$&  25$\times$& 7$\times$\\
         LLM-DA (All)&  105$\times$&  50$\times$&  75$\times$& 30$\times$\\ \bottomrule
    \end{tabular}}
    \caption{Augmentation rate of different methods on different datasets relative to ``Gold'' entries.}
    \label{table:aug-rate}
\end{table}

\begin{table*}[!t]
    \centering
    \resizebox{\textwidth}{!}{
        \begin{tabular}{l|l|cccc|cccc}
        \toprule
        \multirow{2}*{Dataset} &  \multirow{2}*{Method} & \multicolumn{4}{c|}{{BERT-base-cased}} & \multicolumn{4}{c}{{RoBERTa-base}} \\
       ~ & ~ & 5-shot & 10-shot & 15-shot & 20-shot & 5-shot & 10-shot & 15-shot & 20-shot \\
       \midrule
       
        \multirow{9}{*}{{CoNLL'03}} & Gold & 1.58 (0.16) & 3.33 (6.44) & 8.72 (16.13) & 34.39 (23.60) & 3.95 (1.06) & 3.35 (0.79) & 3.45 (0.26) & 33.90 (20.36) \\
        ~ & LSMS & 34.25 (7.80) & 50.06 (3.02) & 57.76 (4.06) & 61.28 (2.73) & 31.14 (11.08) & 47.83 (2.94) & 52.30 (3.99) & 56.94 (3.10) \\ 
        ~ & DAGA & 17.11 (9.37) & 40.22 (7.79) & 39.51 (14.24) & 46.26 (4.82) & 3.04 (4.97) & 8.89 (12.54) & 2.98 (3.39) & 13.45 (11.70) \\
        ~ & MELM & 1.05 (2.16) & 7.59 (12.13) & 35.49 (4.58) & 45.71 (3.41) & 4.35 (0.97) & 1.69 (2.21) & 38.29 (7.36) & 39.99 (5.78) \\
        \rowcolor{gray!10} \cellcolor{gray!0}CoNLL'03 & \cellcolor{gray!10}LLM-DA (Context) & \underline{40.51 (7.46)} & 51.79 (4.74) & 58.87 (4.19) & 62.74 (3.79) & \underline{44.50 (8.39)} & \underline{50.82 (2.20)} & \underline{55.54 (2.83)} & 57.77 (2.72) \\
        \rowcolor{gray!10} \cellcolor{gray!0}~ & LLM-DA (Entity) & \underline{39.53 (9.02)} & \underline{51.48 (3.73)} & 57.57 (3.41) & 62.92 (2.26) & \underline{36.25 (9.59)} & 48.72 (3.70) & \underline{55.49 (3.95)} & 57.92 (3.23) \\
        \rowcolor{gray!10} \cellcolor{gray!0}~ & LLM-DA (Both) & \underline{52.78 (5.99)} & \underline{60.22 (3.05)} & \underline{65.68 (3.52)} & \underline{66.21 (3.25)} & \underline{52.56 (5.38)} & \underline{58.10 (2.93)} & \underline{61.39 (3.81)} & \underline{63.96 (2.80)} \\ 
        \rowcolor{gray!10} \cellcolor{gray!0}~ & LLM-DA (All) & \underline{\textbf{54.04} (5.93)} & \underline{\textbf{61.25} (3.08)} & \underline{\textbf{66.64} (2.70)} & \underline{\textbf{67.68} (2.47)} & \underline{\textbf{54.62} (4.14)} & \underline{\textbf{59.32} (3.56)} & \underline{\textbf{64.05} (1.78)} & \underline{\textbf{64.40} (2.22)} \\
        \midrule
        
        \multirow{9}{*}{{OntoNotes 5.0}} & Gold & 10.34 (7.68) & 38.41 (4.48) & 49.44 (1.52) & 57.56 (3.19) & 0.08 (0.01) & 18.57 (7.89) & 35.12 (10.12) & 49.65 (5.08) \\ 
        ~ & LSMS & 39.90 (3.64) & 53.19 (3.09) & 58.47 (1.64) & \textbf{64.02} (2.66) & 36.96 (6.58) & 47.74 (3.76) & 52.71 (2.82) & \textbf{59.75} (1.45) \\ 
        ~ & DAGA & 6.42 (4.54) & 17.26 (14.19) & 24.62 (18.71) & 19.84 (20.67) & 0.01 (0.04) & 0.16 (0.35) & 6.90 (12.39) & 4.69 (8.72) \\
        ~ & MELM & 0.54 (1.04) & 36.17 (2.75) & 48.25 (3.32) & 55.42 (2.22) & 5.96 (6.60) & 23.11 (8.22) & 38.87 (5.11) & 44.26 (14.91) \\ 
        \rowcolor{gray!10} \cellcolor{gray!0}OntoNotes 5.0 & LLM-DA (Context) & {34.54 (6.01)} & {46.75 (4.39)} & {52.12 (2.91)} & 57.34 (3.24) & {28.07 (4.52)} & {37.84 (4.98)} & {45.02 (4.21)} & 50.90 (3.27) \\ 
        \rowcolor{gray!10} \cellcolor{gray!0}~ & LLM-DA (Entity) & 42.68 (5.32) & \textbf{54.79} (2.89) & \textbf{58.52} (3.34) & 63.62 (1.94) & {27.08 (9.53)} & 45.66 (5.98) & \textbf{54.27} (4.58) & {57.77 (2.54)} \\ 
        \rowcolor{gray!10} \cellcolor{gray!0}~ & LLM-DA (Both) & \underline{44.25 (6.75)} & {50.66 (3.72)} & {52.80 (2.52)} & 55.54 (2.58) & 40.01 (5.49) & 46.12 (3.90) & {49.65 (4.34)} & {54.26 (3.02)} \\ 
        \rowcolor{gray!10} \cellcolor{gray!0}~ & LLM-DA (All) & \underline{\textbf{46.35} (5.01)} & 51.89 (2.99) & {55.74 (3.11)} & {60.05 (2.27)} & \underline{\textbf{43.92} (5.68)} & \underline{\textbf{50.29} (4.01)} & 53.97 (2.74) & {56.97 (2.24)} \\
        \midrule
        
        \multirow{9}{*}{{MIT-Movie}} & Gold & 17.59 (9.38) & 49.17 (2.57) & 55.89 (2.84) & 61.94 (2.06) & 2.88 (6.16) & 49.30 (2.45) & 57.23 (2.58) & 64.50 (1.41) \\
        ~ & LSMS & 45.28 (2.34) & 56.64 (2.43) & 60.01 (2.38) & 65.80 (1.46) & 46.84 (2.87) & 59.78 (2.53) & 63.09 (2.02) & 66.69 (1.46) \\
        ~ & DAGA & 27.68 (5.89) & 39.89 (4.93) & 38.99 (4.17) & 47.61 (3.56) & 3.37 (5.54) & 11.35 (7.16) & 23.75 (9.20) & 44.17 (6.06) \\ 
        ~ & MELM & 37.08 (4.37) & 54.39 (1.76) & 59.57 (1.82) & 64.71 (1.58) & 30.54 (2.68) & 51.87 (2.87) & 58.09 (1.79) & 63.17 (1.43) \\
        \rowcolor{gray!10} \cellcolor{gray!0}MIT-Movie & LLM-DA (Context) & {42.67 (4.05)} & 54.93 (1.95) & 56.62 (2.10) & 61.72 (2.43) & {39.05 (2.28)} & 52.32 (2.66) & {55.41 (1.91)} & {60.23 (2.96)} \\ 
        \rowcolor{gray!10} \cellcolor{gray!0}~ & LLM-DA (Entity) & \underline{49.20 (4.16)} & \underline{62.32 (2.15)} & \underline{63.96 (2.14)} & \underline{\textbf{68.71} (1.01)} & \underline{55.76 (5.21)} & \underline{\textbf{66.86} (1.56)} & \underline{67.13 (1.29)} & \underline{\textbf{70.49} (1.11)} \\ 
        \rowcolor{gray!10} \cellcolor{gray!0}~ & LLM-DA (Both) & \underline{51.08 (4.82)} & \underline{59.61 (1.37)} & 60.87 (2.78) & 66.18 (1.58) & \underline{54.23 (5.57)} & \underline{62.67 (1.90)} & 64.02 (1.54) & 66.85 (1.71) \\ 
        \rowcolor{gray!10} \cellcolor{gray!0}~ & LLM-DA (All) & \underline{\textbf{52.85} (3.70)} & \underline{\textbf{62.80} (1.22)} & \underline{\textbf{64.86} (2.21)} & \underline{68.55 (1.69)} & \underline{\textbf{58.94} (3.99)} & \underline{66.07 (2.01)} & \underline{\textbf{67.65} (1.98)} & \underline{70.24 (1.21)} \\
        \midrule
        
        \multirow{9}{*}{{FEW-NERD}} & Gold & 18.82 (1.28) & 32.23 (1.95) & 39.29 (1.79) & 43.47 (0.80) & 12.45 (1.86) & 29.47 (1.51) & \textbf{38.76} (1.93) & \textbf{43.03} (1.35) \\ 
        ~ & LSMS & 28.00 (2.56) & 35.53 (1.67) & 39.72 (1.84) & 42.71 (1.08) & 24.19 (2.98) & 32.27 (1.72) & 36.59 (2.30) & 40.94 (1.62) \\ 
        ~ & DAGA & 0.15 (0.48) & 20.38 (5.60) & 30.05 (10.82) & 38.01 (8.41) & 0.13 (0.31) & 11.13 (9.16) & 22.59 (12.39) & 30.11 (14.19) \\ 
        ~ & MELM & 22.42 (4.10) & 35.11 (3.50) & 40.71 (1.79) & \textbf{44.36} (2.14) & 19.42 (2.57) & 33.42 (2.30) & 38.66 (2.69) & 41.70 (1.33) \\ 
        \rowcolor{gray!10} \cellcolor{gray!0}FEW-NERD & LLM-DA (Context) & 23.60 (1.81) & {30.02 (2.24)} & 34.56 (1.72) & 37.81 (1.78) & 17.76 (2.34) & {26.00 (2.62)} & {32.74 (1.91)} & 35.66 (1.79) \\ 
        \rowcolor{gray!10} \cellcolor{gray!0}~ & LLM-DA (Entity) & \underline{\textbf{33.98} (2.80)} & \underline{\textbf{39.87} (1.29)} & \textbf{41.67} (2.26) & 43.88 (1.72) & \underline{\textbf{31.89} (1.83)} & \underline{\textbf{36.40} (1.90)} & 37.84 (2.48) & 40.20 (1.50) \\ 
        \rowcolor{gray!10} \cellcolor{gray!0}~ & LLM-DA (Both) & \underline{32.35 (1.45)} & 36.64 (1.68) & 39.35 (1.44) & 41.05 (1.21) & \underline{29.65 (1.62)} & 34.15 (2.42) & 37.44 (1.81) & {39.61 (1.11)} \\ 
        \rowcolor{gray!10} \cellcolor{gray!0}~ & LLM-DA (All) & \underline{32.02 (1.43)} & 36.42 (1.30) & 38.97 (2.67) & 40.42 (1.61) & \underline{29.55 (1.53)} & \underline{34.86 (1.11)} & 36.43 (1.59) & {38.61 (1.95)} \\
        \bottomrule
        
        \end{tabular}}
    \caption{Evaluation results based on micro-F1 scores. ``$k$-shot'' ($k \in \{5, 10, 15, 20\}$) refers to the occurrence of each label type $k$ times. We conduct ten independent replications of all experiments using different random seeds. The mean and standard deviation values are reported. The best results in each setting are highlighted in \textbf{bold}. \underline{Underlined} results indicate significantly better than all baseline methods, determined by a paired Student's t-test with a p-value of less than 0.05.}
    \label{tab:main_result}
\end{table*}

\subsection{Experimental Settings}

Our experiments consist of two phases: data augmentation and model validation.

In the data augmentation phase, we employ gpt-3.5-turbo, developed by OpenAI~\footnote{\url{https://platform.openai.com/docs/models/gpt-3-5-turbo}}, which is the most capable model from the GPT-3.5 sub-series. To fully harness its capabilities, we maximize the model's output length to 2048 tokens. For the sake of reproducibility in our experimental results, we typically set the temperature parameter to 0. However, during the both-level data augmentation phase, we set the temperature parameter to 1 to enhance the diversity of entities.

It's important to note that our augmentation multiplicity varies due to differences in the number of labels and sample sizes across various datasets. Specific augmentation multiplicity values can be found in Table \ref{table:aug-rate}. In cases involving the same dataset, with the exception of certain baseline methods (e.g., MELM) where optimal augmentation rates are specified in the original paper, we adjust the augmentation rates of other methods to match the augmentation ratios of LLM-DA (All). This careful adjustment ensures a fair comparison of experimental results.

During the model validation phase, we merge the augmented data with the original data
using the inside-outside tagging scheme~\cite{IO-scheme}. In the experiments, we set the batch size to 8 and utilize the AdamW~\cite{AdamW} optimizer with a learning rate of 3e-5. The models are trained for 20 epochs, and the best-performing model on the validation set is selected for testing. Meanwhile, we implement an early-stop strategy, terminating training if there is no improvement after 5 epochs.

Furthermore, to mitigate the impact of label noise resulting from the labeling process, we employ the generalized cross-entropy loss~\cite{GCELoss} for  both our approach and all data augmentation baselines:
\begin{equation*}
    \mathrm{GCELoss}_q(\hat{y}, e_j)=\frac{1-\hat{y}_j^q}{q}
\end{equation*}
where $\hat{y}$ represents the predicted probability distribution, $e_j$ represents the actual label as class $j$, $\hat{y_j}$ represents the probability of prediction as class $j$, and $q$ is a hyperparameter that is set to 0.5 uniformly. For baseline ``Gold'', we assume it is free of noise and use cross-entropy loss.

\section{Experimental Analysis}

\subsection{Main Results}
We evaluate our method by comparing it with existing methods across four few-shot settings: 5-shot, 10-shot, 15-shot, and 20-shot. We perform 10 independent repetitions and calculate the mean and standard deviation of the micro-F1 score for each method. The results are summarized in Table \ref{tab:main_result}.
From the experimental results, we have the following observations.

\paragraph{Improvement of Performance}
In scenarios with limited resources, our method consistently enhances model performance when compared to existing approaches. To illustrate, on the CoNLL'03 dataset containing only four entity types, our method exhibits a substantial performance improvement of at least 30 points compared to any other existing method. Likewise, on the FEW-NERD dataset, encompassing 66 fine-grained entity types, our method outperforms the no data augmentation approach by a margin of 5-15 points in the 5/10-shot settings.

\paragraph{Comparison of Strategies}
By synthesizing the insights from Tables \ref{datasets} and \ref{tab:main_result}, we can discern the impact of different levels of data augmentation methods on model performance across various datasets. An intriguing observation comes to light: when dealing with smaller datasets, such as the CoNLL'03 dataset, both-level and context-level augmentations yield more noticeable performance improvements compared to relying on entity-level augmentation.
This observation underscores the effectiveness of context-level augmentation strategies in elevating model performance across the 14 domains under examination. Additionally, both-level augmentation capitalizes on the strengths of both context-level and entity-level augmentation, delivering optimal results in specific scenarios.

However, as dataset size increases, the advantage of context-level augmentation diminishes, while the benefits of entity-level augmentation continue to grow, potentially surpassing the both-level augmentation. We speculate that this shift in performance dynamics can be attributed to the model's increasing focus on fitting entity information rather than contextual information.
Furthermore, context-level augmentation introduces variations from the test set into the augmented data distribution, which become more pronounced with larger datasets, ultimately exerting a negative impact on performance. This drawback also extends to both-level augmentation, which relies on context-level augmented data as a foundation for further variation, exacerbating the negative impact on performance.

\begin{figure}[!t]
    \centering
    \label{fig:chatgpt}
    \includegraphics[width=0.8\linewidth, height=0.5\linewidth]{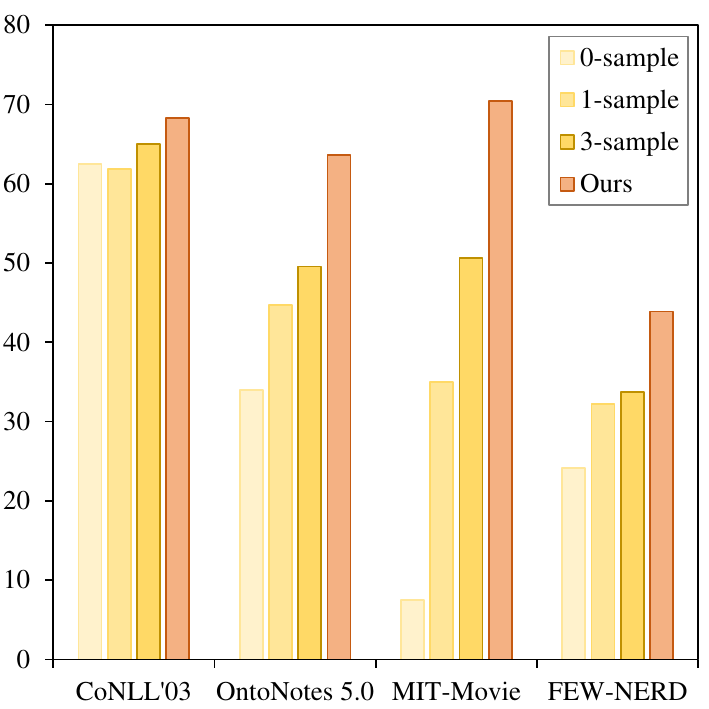}
    \caption{
    Performance of ChatGPT using 0/1/3-sample with that of smaller model trained using LLM-DA data.}
    \label{fig:performace-gpt}
    
\end{figure}

\paragraph{Applicable Scenarios}
Compared to alternative approaches, LLM-DA offers more pronounced advantages in two scenarios. 
On the one hand, it excels in \textbf{extreme low-resource situations}. In the 5-shot scenario, LLM-DA achieves a significantly greater performance boost across all four datasets than other methods. However, it's important to note that this benefit diminishes as the number of examples increases. In particular, when dealing with the FEW-NERD dataset, containing roughly 860 data samples in a 20-shot setting, the use of any data augmentation method can negatively impact model performance. Nevertheless, LLM-DA still maintains a high overall performance relative to its counterparts. This is because, beyond a certain data volume threshold, the adverse effects of noise introduced through data augmentation may outweigh the advantages gained from increased data.
On the other hand, LLM-DA proves highly effective when dealing with \textbf{domain-specific data}. On datasets such as CoNLL'03 and MIT-Movie, whose domains align with our context-level augmentation, the majority of our method's results significantly outperform those of all other methods. However, on datasets like OntoNotes 5.0 and FEW-NERD, while our method generally outperforms others, a few methods manage to achieve comparable results. This discrepancy may stem from the fact that our method imposes certain constraints on the augmented data without automatically adapting to the specific domain corresponding to the data.

\subsection{Performance Analysis of ChatGPT}

To directly assess the LLMs' performance in the NER task, we employ ChatGPT as a representative to evaluate it across four datasets. Following the experimental protocols established in prior studies that evaluated ChatGPT's performance in the NER task~\cite{Stanford,GPT-chen, GPT-ye}, we conduct tests in both zero-shot and few-shot scenarios.~\footnote{The specific prompts can be found in Appendix \ref{appendix:gpt}.} However, owing to ChatGPT's \emph{token length limitations}, we are restricted to incorporating a maximum of three examples simultaneously. The results of our final assessment are depicted in Figure \ref{fig:performace-gpt}.

Upon a comprehensive analysis of the outcomes, several key observations emerge:
1) our method consistently outperforms ChatGPT in the NER task across all four datasets. When we augment the data with LLM-DA, the smaller model consistently surpasses ChatGPT. It's worth noting that ChatGPT shows results closer to our methods only in the case of the CoNLL'03 dataset, which comprises only four labels. This observation can be attributed to the dataset's simplicity and its status as a widely-used public benchmark, raising the possibility that ChatGPT encountered it during its training. Consequently, the practical benefits of our method will become more pronounced;
2) ChatGPT is not suitable for direct application to the NER task, which aligns with the findings in \citet{GPT-ye}. This limitation is particularly pronounced when addressing datasets with a greater variety of labeled types, such as MIT-Movie and FEW-NERD;
and 3) consistent with \citet{in-context}, demonstrations fail to mitigate ChatGPT's performance challenges in the NER task. The contextual constraints of ChatGPT hinder our ability to introduce additional examples, and the performance gains from such additions appear to diminish over time.

\subsection{Linguistic Quality Analysis}

\begin{table}[!t]
    \centering
    \resizebox{\linewidth}{!}{%
    \begin{tabular}{lcccc}
    \toprule
    {Methods} & LSMS & DAGA & MELM & LLM-DA \\
    \midrule
    {Informativeness ($\uparrow$)} & 21.28 & 23.40 & 23.91 & \textbf{24.14}  \\
    {Readability ($\uparrow$)} &    10.29 & 7.47 & 9.07 & \textbf{11.53} \\
    {Grammatical Errors ($\downarrow$)}  & 1.49 & 1.28 & 3.68 & \textbf{0.62} \\
    \bottomrule
    \end{tabular}
    }
    \caption{Linguistic quality of augmented data.}
    \label{informativeness}
\end{table}
As depicted in Figure \ref{fig:augmented-data}, data produced by earlier data augmentation methods frequently exhibit syntactic inaccuracies or semantic inconsistencies. To quantitatively demonstrate the superior quality of the data we have generated, we conduct a comprehensive evaluation, comparing it with data generated by various methods in a 20-shot scenario using the CoNLL'03 dataset.
We analyze informativeness~\cite{Informativeness} using the natural language toolkit~\cite{NLTK}, assess readability using the Flesch-Kincaid method~\cite{flesch-kincaid}, and evaluate grammaticality using the language-check library.

The experimental results presented in Table \ref{informativeness} indicate that, as all methods generate new data based on existing data, the amount of information in the generated data is comparable. However, LLM-DA exhibits a slight advantage over other methods. This advantage stems from its ability to leverage the world knowledge embedded in LLMs.
Regarding readability and grammaticality, LLM-DA demonstrates a significant advantage by leveraging the excellent rewriting capabilities of LLMs themselves. In conclusion, the data generated by LLM-DA consistently outperforms other existing methods at both syntactic and semantic levels, making it more suitable for data augmentation techniques.

\begin{figure}
    \centering
    \subfigure[DAGA]{
    \label{fig:daga-conll}
    \includegraphics[width=0.45\linewidth]{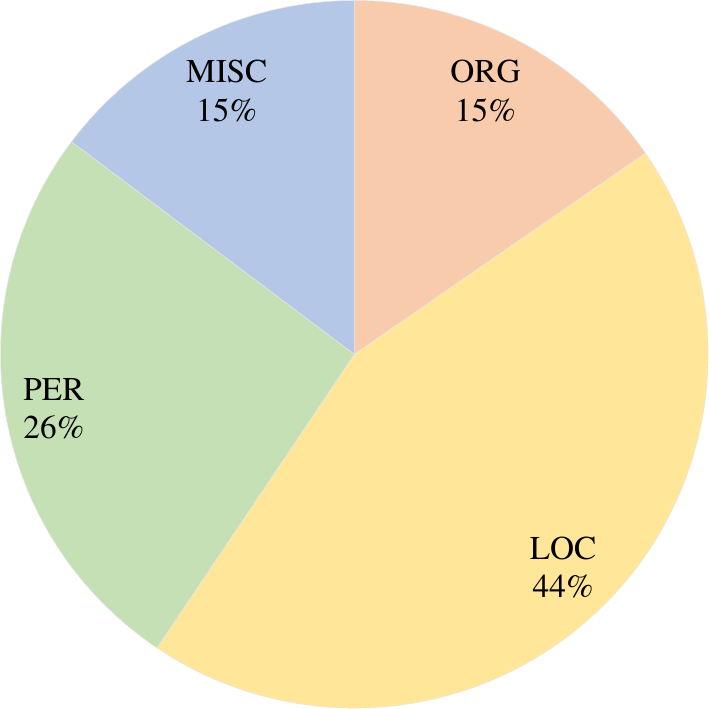}}
    \subfigure[LLM-DA]{
    \label{fig:ours-conll}
    \includegraphics[width=0.45\linewidth]{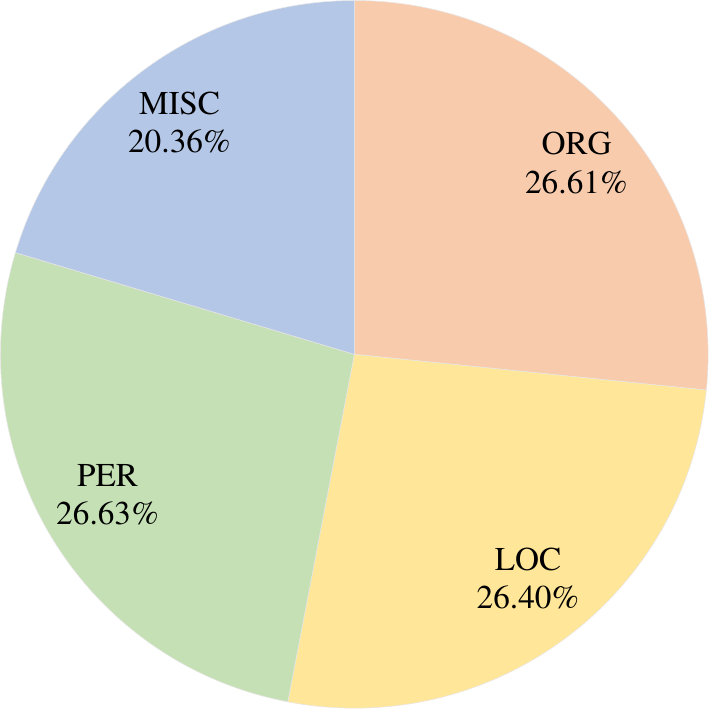}}
    \caption{The entity distribution of data augmented by DAGA (\ref{fig:daga-conll}) and LLM-DA (All) (\ref{fig:ours-conll}).}
    \label{fig:distribution}    
\end{figure}

\subsection{Further Studies}
\paragraph{Entity Distribution}
As mentioned in Section \ref{sec:intro}, we aim to strike a balance between diversity and controllability when generating data using different prompts. 
In comparison, DAGA generates data without any restrictions.
To compare the data generated by DAGA and LLM-DA(All), we conduct a comparison on the CoNLL'03 dataset in a 20-shot scenario. Figure \ref{fig:distribution} clearly shows that the data augmented by DAGA exhibits a severely unbalanced entity distribution due to the absence of restrictions during generation. Conversely, LLM-DA generates data with a more uniform distribution of entities. By taking this aspect into consideration, LLM-DA fosters improved model training.

\begin{figure}[!t]
    \centering    
    \subfigure[OntoNotes 5.0]{
    \label{fig:onto}
    \includegraphics[width=0.47\linewidth]{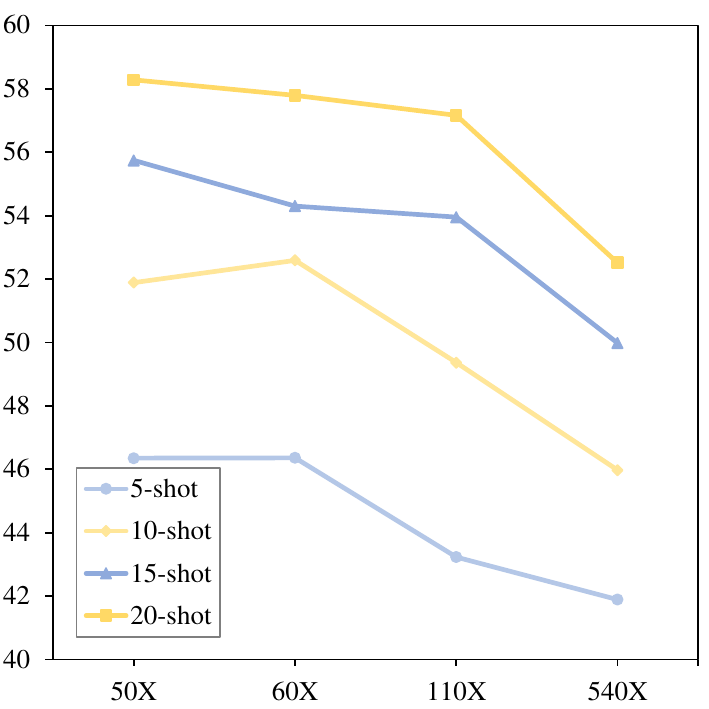}
    }
    \subfigure[FEW-NERD]{
    \label{fig:few-nerd}
    \includegraphics[width=0.47\linewidth]{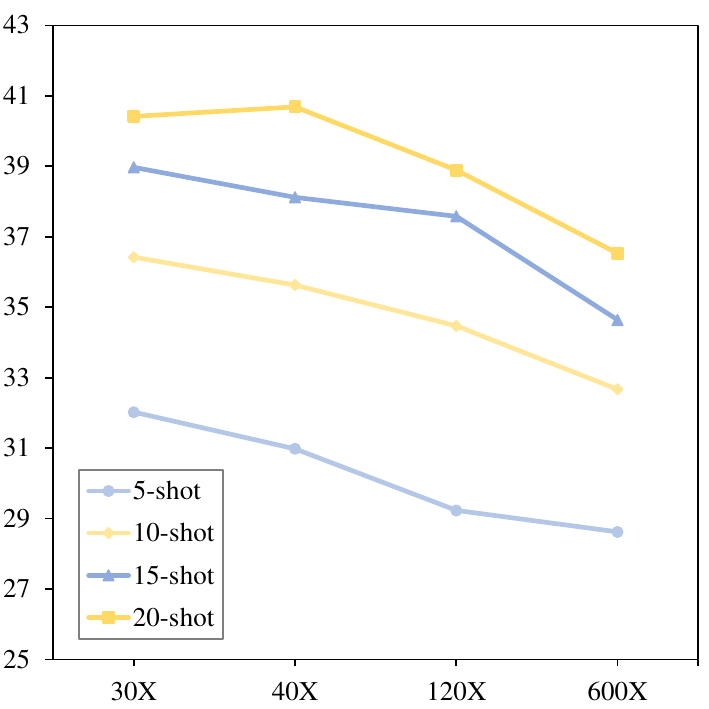}}
    \caption{Performance of BERT on the OntoNotes 5.0 (\ref{fig:onto}) and FEW-NERD (\ref{fig:few-nerd}) datasets when trained with LLM-DA (All) data at various augmentation ratios.}
    \label{fig:aug-ratio}
\end{figure}

\paragraph{Impact of the Number of Augmented Data}
To investigate whether model performance continues to improve with higher augmentation ratios, we extend the augmentation ratio using LLM-DA (All) as a basis and conduct experiments with BERT on the OntoNotes 5.0 and FEW-NERD datasets. The results, as depicted in Figure \ref{fig:aug-ratio}, reveal that further increasing the augmentation ratio does not significantly enhance model performance; in fact, it even lead to a decline in performance.
This phenomenon arises because the augmented data is generated based on the pre-existing sampled data, which inherently has limited diversity, thus constraining the quality of the augmented data. As the augmentation ratio reaches a certain threshold, data diversity reaches its upper limit. Continued expansion of the data exacerbates the disparity between the data distribution and the test data, introducing more noise and thereby impeding the model training process. Consequently, after a certain amount of augmented data is reached, \emph{more is not necessarily better}, and the key to improving model performance lies in enhancing the quality of the data.

\section{Related Work}
\paragraph{Few-Shot NER Task}
Due to the resource-intensive nature of labeling NER data, it is more practical to address the few-shot NER task~\cite{huang2020few}. While the current main solutions for few-shot NER primarily focus on model-level improvements, such as using prototype networks~\cite{fritzler2019few} or template-based models~\cite{template_based,template_free}, our work differentiates itself by focusing on the data level. Our objective is to tackle the fundamental challenge of insufficient annotation data for NER in few-shot scenarios.

\paragraph{Data Augmentation for the NER Task}
Data augmentation methods for NER can be categorized into sentence transformations~\cite{EDA, LSMS} and methods based on model generation~\cite{DAGA, MELM}. However, these methods exhibit inherent limitations. They have the potential to break the syntactic structure of sentences or generate sentences that do not align with contextual semantics, leading to excessive noise in model training.
In our approach, we leverage LLMs to generate a large quantity of diverse and high-quality new data for NER. Notably, our method preserves the grammatical and semantic structure of the sentences, ensuring more effective and accurate training.

\paragraph{LLMs}
LLMs, trained on extensive text corpora, demonstrate remarkable proficiency in acquiring vast world knowledge~\cite{GPT,instructGPT, BLOOM, PaLM, LLaMA, Llama-2}. They have been successfully employed for data augmentation in certain straightforward tasks that merely involve direct rewriting~\cite{LAMBADA, DA-intent, Auggpt}. However, LLMs already exhibit high performance in these tasks, rendering further augmentation unnecessary.
Our approach specifically targets the NER task, which highlights the performance limitations of LLMs~\cite{GPT-chen, GPT-ye}. By carefully considering the inherent constraints of LLMs and the distinct characteristics of NER, we strive to enhance the utilization of LLMs in NER task.

\section{Conclusion}
In this paper, we introduce LLM-DA, a data augmentation method utilizing LLMs for the few-shot NER task. Our method harnesses the rewriting capabilities and vast world knowledge of LLMs to effectively address syntactic or semantic inconsistencies encountered in existing approaches. Additionally, we strike a careful balance between data diversity and unpredictability by incorporating prompts at both the context and entity levels. Through extensive experiments, LLM-DA demonstrates performance superiority or parity with current methods, making a valuable contribution to enhancing model performance.


\section*{Limitations}
While we have made efforts to account for a wide range of scenarios, our approach also has two limitations. First, contextual rewriting in our approach is constrained by design strategies, without adapting the model to specific sentence domains. Nonetheless, we have mitigated this limitation to some extent by considering common domains extensively. Second, LLMs have a token length restriction, which limits the diversity of replacement entries for entities. We believe that the strengths of our approach can be further emphasized by expanding the generation of entities.

\section*{Ethics Statement}
Our approach leverages LLMs to generate data that is independent of existing training data, instead drawing on the model's knowledge acquired from extensive pre-trained data. However, it is important to note that the generated data may inherit societal biases present in the pretraining corpus. Therefore, we recommend performing human inspection of the generated data to mitigate the risk of propagating such biases to downstream classification models in practical applications.

\bibliography{custom}

\appendix

\section{Prompts for LLM-DA}
\label{appendix:prompts}

We design a total of 17 augmentation prompts, consisting of 14 prompts for context-level augmentation (Table~\ref{tab:prompts-context}), one prompt for entity-level augmentation (Table \ref{tab:prompts-entity}), one prompt for both-level augmentation (Table \ref{tab:prompts-both}), and one prompt for noise injection (Table \ref{tab:prompts-noise}).

\section{Prompts for Processing the NER Task}
\label{appendix:gpt}

To assess ChatGPT's performance on NER tasks, we devise specific prompts for direct utilization of gpt-3.5-turbo in tackling these tasks, as outlined in Table \ref{tab:prompts-gpt}.

\begin{table}[h]
\centering
    \begin{tabular}{m{0.92\linewidth}}
    \toprule
      Given a sentence: """ \\
\{sentence\}\\
"""\\
\\
Given entities in above sentence separated by comma: """\\
\{entities\}\\
"""\\
\\
Please generate 5 new sentences using \{new context\}. In each new sentence, use each given entity exactly once, keep their entity type in the given sentence, introduce no other entities.\\
\\
Desired format:\\
Kept Entities: <all kept entities separated by comma>\\
New sentence: <new sentence>\\
\\
Kept Entities: <all kept entities separated by comma>\\
New sentence: <new sentence>
\\ \bottomrule
\end{tabular}
\caption{Prompts designed for context-level augmentation. Replace ``\{sentence\}'' with the original sentence. Replace ``\{entities\}'' with the entities from the original sentence. Replace ``\{new context\}'' with the strategies to rephrase the context.}
\label{tab:prompts-context}
\end{table}

\begin{table}[ht]
\centering
    \begin{tabular}{m{0.92\linewidth}}
    \toprule

Given a sentence: """\\
\{sentence\}\\
"""\\
\\
Given entities in above sentence separated by comma: """\\
\{entities\}\\
"""\\
\\
Please generate 20 sentences, replace given entities with new entities of the same type, keep other words.\\
\\
Desired format:\\
Replaced Entities: <given entity -> new entity, given entity -> new entity>\\
New sentence: <new sentence>\\
\\
Replaced Entities: <given entity -> new entity, given entity -> new entity>\\
New sentence: <new sentence>\\ 
\bottomrule
\end{tabular}
\caption{Prompts designed for entity-level augmentation. Replace ``\{sentence\}'' with the original sentence. Replace ``\{entities\}'' with the entities from the original sentence.}
\label{tab:prompts-entity}
\end{table}

\begin{table}[ht]
\centering
    \begin{tabular}{m{0.92\linewidth}}
    \toprule
Given a sentence: """\\
\{sentence\}\\
"""\\
\\
Given entities in above sentence separated by comma: """\\
\{entities\}\\
"""\\
\\
Please replace given entities with new entities of the same type, keep other words.\\
\\
Desired format:\\
Replaced Entities: <given entity -> new entity, given entity -> new entity>\\
New sentence: <new sentence>\\ \bottomrule
\end{tabular}
\caption{Prompts designed for both-level augmentation. Replace ``\{sentence\}'' with the original sentence. Replace ``\{entities\}'' with the entities from the original sentence.}
\label{tab:prompts-both}
\end{table}

\begin{table}[ht]
\centering
    \begin{tabular}{m{0.92\linewidth}}
    \toprule
 Given a sentence: """\\
\{sentence\}\\
"""\\
\\
Given entities in above sentence separated by comma: """\\
\{entities\}\\
"""\\
\\
Please use some common spelling mistakes randomly in the given sentence.\\
\\
Desired format:\\
Replaced Entities: <given entity -> new entity, given entity -> new entity>\\
New sentence: <new sentence>\\
\bottomrule
\end{tabular}
\caption{Prompts designed for noise injection. Replace ``\{sentence\}'' with the original sentence. Replace ``\{entities\}'' with the entities from the original sentence.}
\label{tab:prompts-noise}
\end{table}

\begin{table}[ht]
    \centering
    \begin{tabular}{m{0.17\linewidth}|m{0.7\linewidth}}
    \toprule
    \# Shot & \multicolumn{1}{c}{Prompt} \\ \midrule
    \multirow{6}*{zero-shot} & Please identify \{entity\_categories\} Entity from the given text.\\
    ~ & ~\\
    ~ & Desired format: \{format\}\\
    ~ & ~\\
    ~ & Text: \{sentence\}\\
    ~ & Entity:\\ \midrule

    \multirow{6}*{few-shot} & Please identify \{entity\_categories\} Entity from the given text.\\
    ~ & ~\\
    ~ & \{examples\}\\
    ~ & ~\\
    ~ & Text: \{sentence\}\\
    ~ & Entity:\\
    \bottomrule
    \end{tabular}
    \caption{Prompts designed for gpt-3.5-turbo to do the NER task directly. Replace ``\{entity\_categories\}'' with the entity categories in the dataset. Replace ``\{sentence\}'' with the sentence to be identified. Replace ``\{format\}'' with the desired output format in the zero-shot setting. Replace ``\{examples\}'' with examples in the few-shot setting.}
    \label{tab:prompts-gpt}
\end{table}


\end{document}